\documentclass{article}

\usepackage[nonatbib, final]{nips_2016}

\usepackage[utf8]{inputenc} 
\usepackage[T1]{fontenc}    
\usepackage{hyperref}       
\usepackage{url}            
\usepackage{booktabs}       
\usepackage{amsfonts}       
\usepackage{nicefrac}       
\usepackage{microtype}      

\usepackage[normalem]{ulem}
\usepackage{bm}
\usepackage{listings}
\usepackage{graphicx}
\usepackage{caption}
\usepackage{subcaption}
\usepackage{multirow}
\usepackage{url}
\usepackage{amsmath}

\newcommand{\smalltitle}[1]{\noindent \textbf{\emph{#1}}\enspace}

\title{Tuning the Scheduling of Distributed Stochastic Gradient Descent with Bayesian Optimization}

\author{
  Valentin Dalibard \\
  Computer Laboratory\\
  University of Cambridge\\
  \And
  Michael Schaarschmidt \\
  Computer Laboratory\\
  University of Cambridge\\
  \And
  Eiko Yoneki \\
  Computer Laboratory\\
  University of Cambridge\\
}

\begin{document}

\maketitle

\begin{abstract}
We present an optimizer which uses Bayesian optimization to tune the system parameters of distributed stochastic gradient descent (SGD). Given a specific context, our goal is to quickly find efficient configurations which appropriately balance the load between the available machines to minimize the average SGD iteration time. Our experiments consider setups with over thirty parameters. Traditional Bayesian optimization, which uses a Gaussian process as its model, is not well suited to such high dimensional domains. To reduce convergence time, we exploit the available structure. We design a probabilistic model which simulates the behavior of distributed SGD and use it within Bayesian optimization. Our model can exploit many runtime measurements for inference per evaluation of the objective function. Our experiments show that our resulting optimizer converges to efficient configurations within ten iterations, the optimized configurations outperform those found by generic optimizer in thirty iterations by up to 2$\times$.
\end{abstract}

\section{Introduction}

When implemented on a distributed cluster, neural network training is typically performed via stochastic gradient descent (SGD) using the parameter server architecture \cite{dean2012large}. Some of the available machines are used as \emph{workers} and compute gradient estimates from a small set of training samples each iterations. A possibly overlapping set of machines are used as \emph{parameter servers}. Every iteration, they aggregate the workers' gradient estimates, update the parameters accordingly and send these values back to the workers. Manually tuning the implementation of distributed SGD is a complex task. Workers with more computational power should be assigned a greater fraction of the batch size~-- the total number of training samples processed per iteration. Furthermore, the synchronization cost incurred each iteration grows with the number of workers. Hence, in some cases, it is beneficial to only use a subset of the available machines as workers.

This paper tackles the automated tuning of distributed SGD. Given a specific context, our goal is to find a placement of parameter servers and workers, along with a load balance among workers, that minimizes the average iteration time of SGD. In our experiments, we optimize this schedule over 10 machines of varying computational power leading to over 30 parameters being tuned. 

This high dimensionality means that traditional Bayesian optimization is not directly applicable. To remedy this, we extract some of the structure of the problem to help the optimization converge. We design a probabilistic model of the behavior of distributed SGD and use it within a Bayesian optimization. Our model predicts how long each worker machine will take to perform its assigned load. Furthermore, it predicts the synchronization cost based on the placement of parameter servers and workers, and an inferred bandwidth parameter for each machine. Thanks to its structure, our model is able to leverage many runtime measurements per iteration, such as the computation time of each worker. As a result, it quickly converges towards the true behavior of the computation.


\section{Optimization problem}
\label{sec:optprob}
\smalltitle{Optimization domain.} We tuned the scheduling of a parameter server architecture, implemented in TensorFlow \cite{tensorflow}, to minimize the average iteration time. Given a set of machines, a neural network architecture and a batch size, we set:

\begin{table}[t]
{\small
\begin{center}
\begin{tabular}{| c | c | c | c | c | c |}
\hline
\multirow{2}{*}{Instance Type} & \multirow{2}{*}{\# Hyperthread} & \multirow{2}{*}{GPU} & \multicolumn{3}{| c |}{\# per setting}  \\
 & & & A & B & C \\\hline
\texttt{g2.2xlarge} & 8 & 1 K520 & 0 & 1 & 2 \\ \hline
\texttt{c4.2xlarge} & 8 & / & 6 & 3 & 2 \\ \hline
\texttt{c4.4xlarge} & 16 & / & 2 & 3 & 2 \\ \hline
\texttt{c4.8xlarge} & 36 & / & 2 & 3 & 4 \\ \hline
\multicolumn{3}{| c |}{Total} & 10 & 10 & 10 \\ \hline
\end{tabular}
\caption {Machine and setting specifications. \vspace{-0mm}}
\label{ec2_specs}
\end{center}
}
\end{table}

\begin{table}[t]
{\small
\begin{center}
\begin{tabular}{| c | c | c | c | c | c |}
\hline
Neural Network name & Input Type & Network Type & Size (MB) & Ops (Millions) & Batch size range \\ \hline
GoogleNet \cite{googlenet} & Image & Convolutional & 26.7 & 1582 & $2^6 - 2^9$  \\ \hline
AlexNet \cite{alexnet} & Image & Convolutional & 233 & 714 & $2^8 - 2^{11}$ \\ \hline
SpeechNet \cite{speechnet} & Audio & Perceptron & 173 & 45.3 & $2^{13} - 2^{16}$ \\ \hline
\end{tabular}
\vspace{1mm}
\caption {The three neural networks used in our experiments. The name ``SpeechNet'' is introduced by us for clarity, this network was recently proposed for benchmarking~\cite{deepmark}. \vspace{-0mm}}
\label{net_specs}
\end{center}
}
\end{table}

\begin{itemize}
 \item Which subset of machines should be used as workers.
 \item Which (possibly overlapping) subset of machines should be used as parameter servers.
 \item Among working machines, how to partition the workload. For machines with GPUs, this includes the partition of the workload between their CPU and GPUs. Each device gets assigned a number of inputs to process per iteration. The total number of inputs must sum up to the batch size.
\end{itemize}
There are effectively two boolean configuration parameters per machine setting whether it should be a worker and/or a parameter server and one to two integer parameters per machine, depending on whether it has a GPU, specifying the load. In our experiments, we tune the scheduling in the three Amazon EC2 settings described in Table \ref{ec2_specs}, designed to recreate heterogeneous settings. Each contains 10 machines of varying computational power, leading to 30-32 parameters being optimized. 

\smalltitle{Objective function.} To measure the performance of a configuration, we performed 20 SGD iterations. 
The first few iterations often showed a high variance in performance and hence we report the average time of the last 10 iterations. We found this was enough to get accurate measurements, repeating configurations showed little underlying noise. 

In our experiments, we optimized the distributed training of the three neural networks referenced in Table \ref{net_specs} with various batch sizes. The batch sizes for each network were selected to explore the tradeoff with processing speed as lower batch sizes tend to improve final result accuracy \cite{1609.04836} at the cost of less parallelism.

\begin{figure*}[t]
    \centering
     \includegraphics[width=\textwidth]{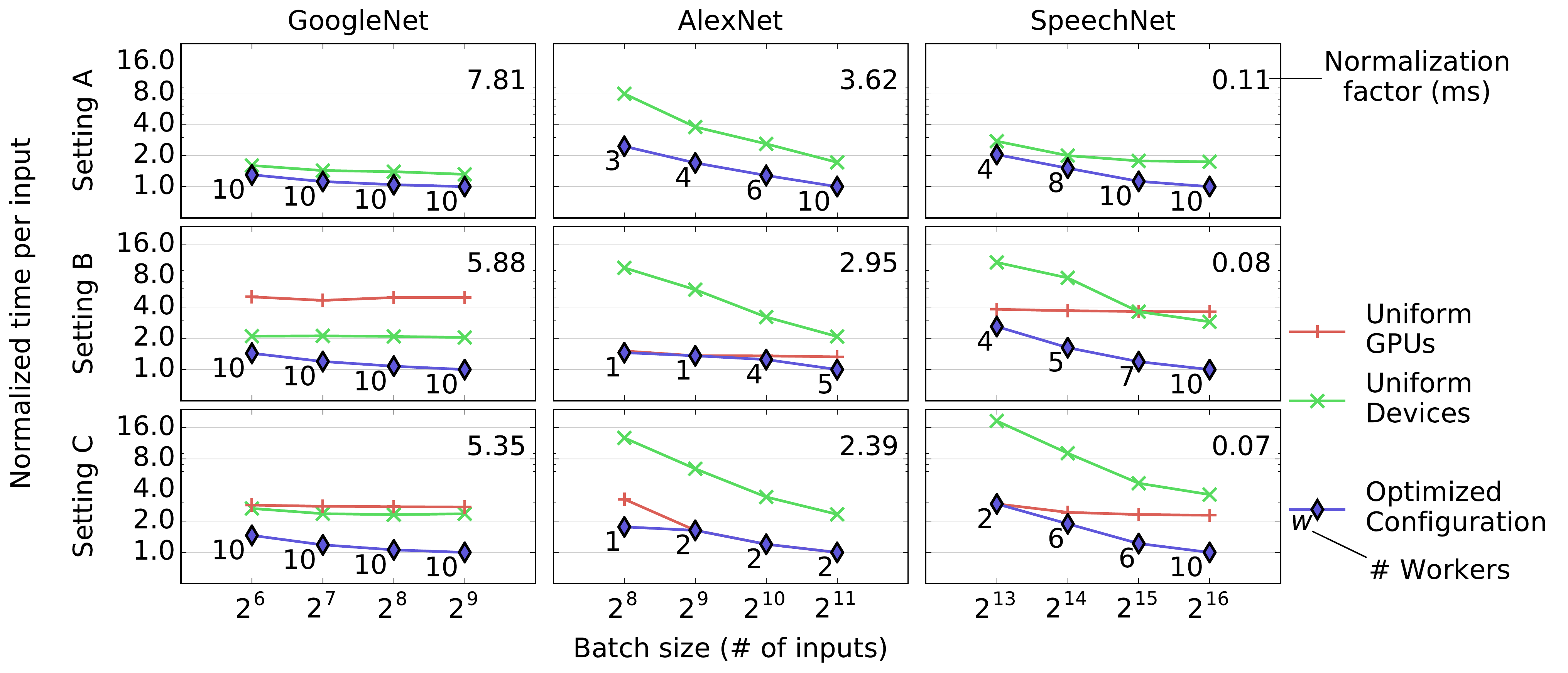}
  \caption{Normalized time per input (lower is better) of simple and optimized configurations on each experiment. Within each sub-graph, results are normalized by the best achieved time per input. This is always the one of the optimized configuration on the largest batch size (the lower right point of each sub-graph). The normalization factor, i.e. the best time per input, is shown at the top right of each sub-graph in milliseconds. For each optimized configuration, we report the number of workers used. }
  \label{fig:simple-conf}
\end{figure*}

\section{A bespoke optimizer for distributed SGD}
\label{sec:pm}
We constructed a \emph{bespoke optimizer} which exploits the known structure of the objective function to converge within few iterations. The optimizer uses Bayesian optimization with a probabilistic model of the behavior of distributed SGD. The model contains several independent parts.

\smalltitle{Individual device computation time.} For each device -- CPU or GPU -- on a worker machine, we modeled the time needed to perform its assigned workload. This time should be near linear with respect to the number of inputs. However, more inputs increases parallelism and sometimes improves efficiency. Our model fits one Gaussian process (GP) per type of device available (e.g. \texttt{c4.4xlarge} CPU, or Nvidia GPU K520) modeling the rate at which inputs are processed as a function of the number of inputs.

\smalltitle{Individual machine computation time.} For machines with multiple devices, the gradient estimates are summed locally on the CPU before being sent to the parameter servers. This results in a non-negligible aggregation time. For each machine type with multiple devices, we used a GP to model the difference between the maximum computation time of its devices and the total computation time. The GP takes as input the number of inputs allocated to each of its devices to predict the aggregation time. 

\smalltitle{Communication time.} We modeled the communication time as a semi-parametric model. Our parametric model infers a $\mathit{connection\_speed}$ parameter per type of machine. It predicts the total communication time as

$$
  \max_{m \,\in\, \mathit{machines}}\; \frac{\mathit{transfer}(m)}{\mathit{connection\_speed}_m }
$$

\noindent where $\mathit{transfer}(m)$ is the amount of data that must be transfered each iteration by machine $m$. It is a function of whether $m$ is a worker, the number of other workers, and the size of the parameters $m$ holds as a parameter server if any. A GP models the difference between this parametric model and observed communication times. We fit a single communication time model for the entire cluster. Finally, we predict the total time of an SGD iteration as the sum of the maximum predicted individual machine time and the communication time. 

\smalltitle{Bespoke models within Bayesian optimization.} We use the expected improvement acquisition function to select the next configuration to evaluate. The expected improvement of a configuration is estimated via forward sampling. When measuring the iteration time of distributed SGD in the objective function, we also measure the time needed by all devices and machines to perform their assigned workload. This allows us to exploit conditional independence and perform inference on each part of the above model separately. 

The optimizer was implemented on top of our Bayesian optimization framework. While a full overview of the framework is out of the scope of this paper, we want to highlight two of its key components:
\begin{itemize}
 \item A probabilistic programming library which can be used to declare a probabilistic model of the function being optimized. It uses sequential Monte Carlo (SMC) inference as suggested by Wood et al. \cite{wood2014new}. Our library is also able to exploit the conditional independence in the model to perform inference.
 \item An optimization decomposition library which can be used in the numerical optimization stage of Bayesian optimization. In high dimensional domains, off-the-shelf optimization algorithms such as DIRECT \cite{jones1993lipschitzian} or CMA-ES \cite{hansen2001completely} can fail to converge to good values. Our decomposition library allows us to guide the numerical optimization to high performance regions. We use it in the context of distributed SGD to optimize the load associated to each worker machine individually.
\end{itemize}

\section{Experiment results}
\label{sec:eval}



\smalltitle{Comparison with simple configurations.} To show the importance of tuning, we compared our optimized configurations with two simple configurations 1) \emph{Uniform Devices}: a load balanced equally among all devices, and 2) \emph{Uniform GPUs}: a load balanced equally among GPUs (in Settings B and C). In both cases, we set worker machines to also be parameter servers which tends to yield good results. Figure \ref{fig:simple-conf} shows the outcome of each experiment. Our optimized configurations significantly outperform these simple configurations on most experiments.

\begin{figure}[t]
\centering
  \includegraphics[width=0.5\textwidth]{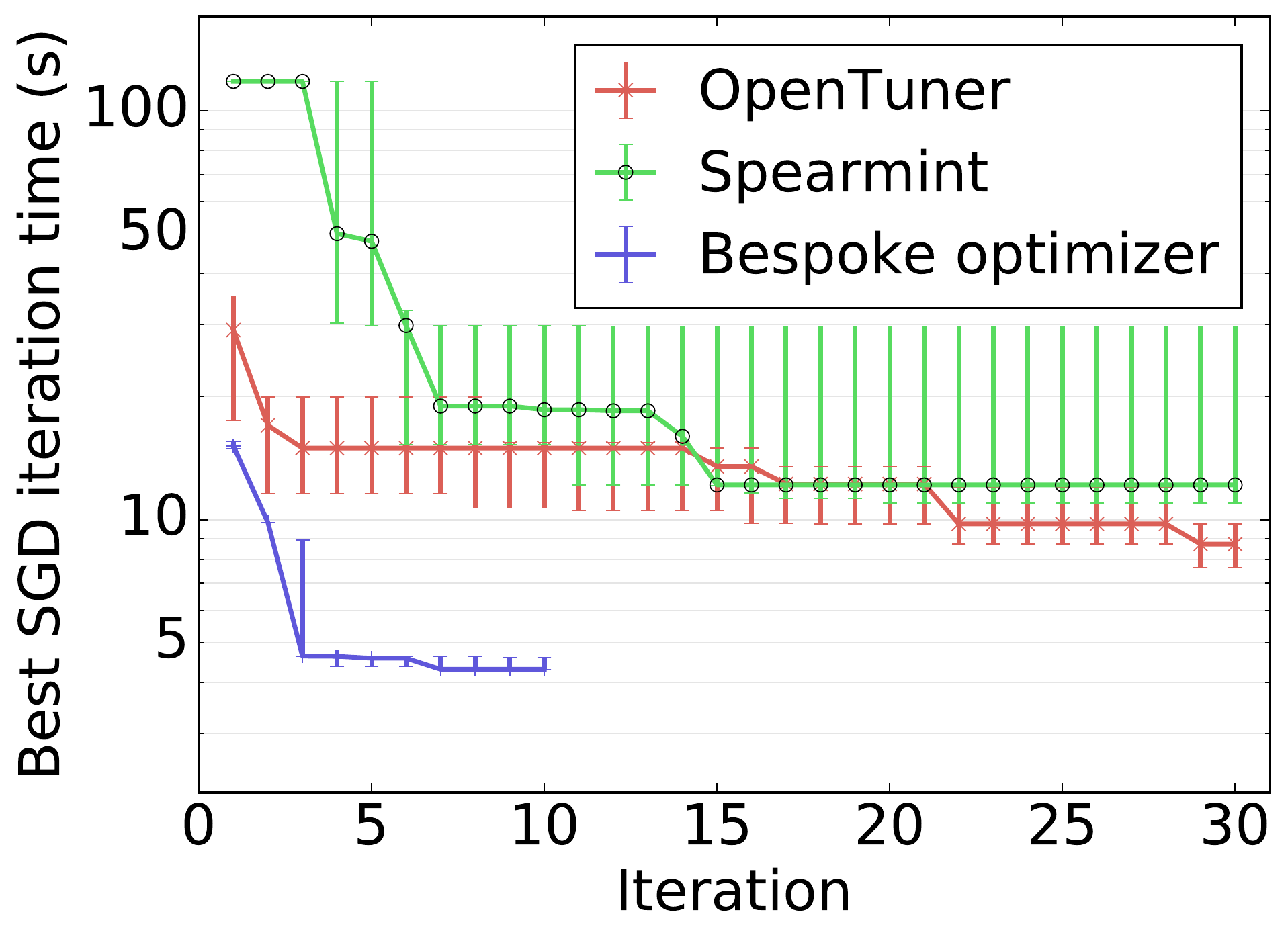}
  \caption{Convergence of the optimizers on Setting C using SpeechNet with a $2^{16}$ batch size.}
  \label{fig:trad-vs-struct}
\end{figure}

\smalltitle{Comparison with traditional Bayesian optimization.} We now consider whether these results could have been achieved with an off-the-shelf optimization tool. Figure \ref{fig:trad-vs-struct} compares the performance of our bespoke optimizer with OpenTuner~\cite{opentuner} and Spearmint~\cite{snoekpractical}, which were each ran for thirty iterations. Each optimization was run three times, we report the best utility found so far of each run at every iteration.
Our bespoke optimizer significantly outperforms generic optimizers. The median best configuration achieved by OpenTuner is 8.71s per SGD iteration, more than twice slower than our median time (4.31s), and not much faster than the Uniform GPUs configuration (9.82s). 

The reason this tuning task is difficult is because the space of efficient configurations is extremely narrow, assigning one of the workers too much work creates a bottleneck, yielding poor performance. By exploiting the known structure of the computation, our bespoke optimizer can discard these regions of low performance and focus on promising configurations.

All of our experiments finished the ten iterations within two hours. As neural networks training typically lasts over a week, the performance gains largely outweigh the tuning overhead, making our optimizer practical in realistic settings. 
\section{Conclusion}
We presented an optimizer which combines Bayesian optimization with a structured probabilistic model to tune the system parameters of distributed SGD. We showed that, by exploiting the known structure of the objective function, our optimizer converged rapidly (10 iterations) and consistently~(stable results over 36 experiments). These results illustrate that Bayesian optimization provides a powerful framework to build bespoke optimizer with high performance on specific problems. Future work could investigate how using such structured probabilistic models could enable the use of experiments other than the objective function within Bayesian optimization.

\bibliographystyle{plain}
\bibliography{paper}

\begin{thebibliography}{10}

\bibitem{tensorflow}
Mart\'{\i}n Abadi, Ashish Agarwal, Paul Barham, Eugene Brevdo, Zhifeng Chen,
  Craig Citro, Greg~S. Corrado, Andy Davis, Jeffrey Dean, Matthieu Devin,
  Sanjay Ghemawat, Ian Goodfellow, Andrew Harp, Geoffrey Irving, Michael Isard,
  Yangqing Jia, Rafal Jozefowicz, Lukasz Kaiser, Manjunath Kudlur, Josh
  Levenberg, Dan Man\'{e}, Rajat Monga, Sherry Moore, Derek Murray, Chris Olah,
  Mike Schuster, Jonathon Shlens, Benoit Steiner, Ilya Sutskever, Kunal Talwar,
  Paul Tucker, Vincent Vanhoucke, Vijay Vasudevan, Fernanda Vi\'{e}gas, Oriol
  Vinyals, Pete Warden, Martin Wattenberg, Martin Wicke, Yuan Yu, and Xiaoqiang
  Zheng.
\newblock {TensorFlow}: Large-scale machine learning on heterogeneous systems,
  2015.
\newblock Software available from tensorflow.org.

\bibitem{opentuner}
Jason Ansel, Shoaib Kamil, Kalyan Veeramachaneni, Jonathan Ragan-Kelley,
  Jeffrey Bosboom, Una-May O'Reilly, and Saman Amarasinghe.
\newblock Opentuner: an extensible framework for program autotuning.
\newblock In {\em Proceedings of the 23rd international conference on Parallel
  architectures and compilation}, pages 303--316. ACM, 2014.

\bibitem{deepmark}
Soumith Chintala.
\newblock Deepmark benchmark.
\newblock \url{https://github.com/DeepMark/deepmark}.

\bibitem{dean2012large}
Jeffrey Dean, Greg Corrado, Rajat Monga, Kai Chen, Matthieu Devin, Quoc~V. Le,
  Mark~Z. Mao, Marc'Aurelio Ranzato, Andrew~W. Senior, Paul~A. Tucker, Ke~Yang,
  and Andrew~Y. Ng.
\newblock Large scale distributed deep networks.
\newblock In {\em Advances in Neural Information Processing Systems 25: 26th
  Annual Conference on Neural Information Processing Systems 2012. Proceedings
  of a meeting held December 3-6, 2012, Lake Tahoe, Nevada, United States.},
  pages 1232--1240, 2012.

\bibitem{hansen2001completely}
Nikolaus Hansen and Andreas Ostermeier.
\newblock Completely derandomized self-adaptation in evolution strategies.
\newblock {\em Evolutionary computation}, 9(2):159--195, 2001.

\bibitem{jones1993lipschitzian}
Donald~R Jones, Cary~D Perttunen, and Bruce~E Stuckman.
\newblock Lipschitzian optimization without the lipschitz constant.
\newblock {\em Journal of Optimization Theory and Applications},
  79(1):157--181, 1993.

\bibitem{1609.04836}
Nitish~Shirish Keskar, Dheevatsa Mudigere, Jorge Nocedal, Mikhail Smelyanskiy,
  and Ping Tak~Peter Tang.
\newblock On large-batch training for deep learning: Generalization gap and
  sharp minima, 2016.

\bibitem{alexnet}
Alex Krizhevsky.
\newblock One weird trick for parallelizing convolutional neural networks.
\newblock {\em arXiv preprint arXiv:1404.5997}, 2014.

\bibitem{speechnet}
Frank Seide, Hao Fu, Jasha Droppo, Gang Li, and Dong Yu.
\newblock 1-bit stochastic gradient descent and its application to
  data-parallel distributed training of speech dnns.
\newblock In {\em INTERSPEECH}, pages 1058--1062, 2014.

\bibitem{snoekpractical}
Jasper Snoek, Hugo Larochelle, and Ryan~Prescott Adams.
\newblock Practical bayesian optimization of machine learning algorithms.
\newblock In {\em Neural Information Processing Systems}, 2012.

\bibitem{googlenet}
Christian Szegedy, Wei Liu, Yangqing Jia, Pierre Sermanet, Scott Reed, Dragomir
  Anguelov, Dumitru Erhan, Vincent Vanhoucke, and Andrew Rabinovich.
\newblock Going deeper with convolutions.
\newblock In {\em Proceedings of the IEEE Conference on Computer Vision and
  Pattern Recognition}, pages 1--9, 2015.

\bibitem{wood2014new}
Frank Wood, Jan-Willem van~de Meent, and Vikash Mansinghka.
\newblock A new approach to probabilistic programming inference.
\newblock In {\em AISTATS}, pages 1024--1032, 2014.

\end{thebibliography}
\end{document}